\documentclass[10pt, twocolumn]{article}
\usepackage{times}
\usepackage{epsfig}
\usepackage{graphicx}
\usepackage{subfigure}
\usepackage{multirow}
\usepackage{paralist}
\usepackage[american]{babel}
\usepackage{graphicx}
\usepackage{amsmath,amssymb} % define this before the line numbering.
\usepackage{color}
\usepackage{authblk}
\usepackage[left=2.5cm, right=2.5cm,top=3cm,bottom=3cm]{geometry}

\title{\textbf{Exploiting Spatial-Temporal Modelling and Multi-Modal Fusion for Human Action Recognition}} % Replace with your title
\author{Dongliang He\thanks{Corresponding author: hedongliang01@baidu.com}, ~Fu Li, ~Qijie Zhao, ~Xiang Long, ~Yi Fu, ~Shilei Wen}
\affil{Baidu Research}
\date{}

\begin{document}
\maketitle

\begin{abstract}
In this report, our approach to tackling the task of ActivityNet 2018 Kinetics-600 challenge is described in detail. Though spatial-temporal modelling methods, which adopt either such end-to-end framework as I3D \cite{i3d} or two-stage frameworks (i.e., CNN+RNN), have been proposed in existing state-of-the-arts for this task, video modelling is far from being well solved. In this challenge, we propose \emph{spatial-temporal network} (StNet) for better joint spatial-temporal modelling and comprehensively video understanding. Besides, given that multi-modal information is contained in video source, we manage to integrate both early-fusion and later-fusion strategy of multi-modal information via our proposed \emph{improved temporal Xception network} (iTXN) for video understanding. Our StNet RGB single model achieves 78.99\% top-1 precision in the Kinetics-600 validation set and that of our improved temporal Xception network which integrates RGB, flow and audio modalities is up to 82.35\%. After model ensemble, we achieve top-1 precision as high as 85.0\% on the validation set and rank No.1 among all submissions.
\end{abstract}

\section{Introduction}
The main challenge lies in extracting discriminative spatial-temporal descriptors from video sources for human action recognition task. CNN+RNN architecture for video sequence modelling \cite{LRCN,yue2015beyond} and purely ConvNet-based video recognition \cite{two-stream,two-stream-stresnet,T-Resnet,tsn,c3d,i3d,p3d} are two major research directions. Despite considerable progress has been made since several years ago, action recognition from video is far from being well solved.

For the CNN+RNN solutions, the feed-forward CNN part is used for spatial modelling, meanwhile the temporal modelling part, e.g., LSTM \cite{lstm} or GRU \cite{gru}, makes end-to-end optimization more difficult due to its recurrent architecture. Taking feature sequence extracted from a video as input, there are many other sequence modelling frameworks or feature encoding methods aiming at better temporal coding for video classification. In \cite{kinetics400_win}, fast-forward LSTM (FF-LSTM) and temporal Xception network are proposed for effective sequence modelling and considerable performance gain is observed against traditional RNN models in terms of video recognition accuracy. NetVLAD \cite{netvlad}, ActionVLAD \cite{actionvlad} and Attention Clusters \cite{attentioncluster} are recently proposed to integrate local features for action recognition and good results are achieved by these encoding methods. Nevertheless, separately training CNN and RNN parts is harmful for integrated spatial-temporal representation learning.

ConvNets-based solutions for action recognition can be generally categorized into 2D ConvNet and 3D ConvNet. Among these solutions, 2D or 3D two-stream architectures achieve state-of-the-art recognition performance. 2D two-stream architectures \cite{two-stream,tsn} extract classification scores from evenly sampled RGB frames and optical flow fields. Final prediction is obtained by simply averaging the classification scores. In this way, temporal dynamics are barely explored due to poor temporal modelling. As a remedy for the aforementioned drawback, multiple 3D ConvNet models are invented for end-to-end spatial-temporal modelling such as T-ResNet \cite{T-Resnet}, P3D \cite{p3d}, ECO \cite{eco}, ARTNet \cite{ARN} and S3D \cite{s3d}. Among these 3D ConvNet frameworks, state-of-the-art solution is non-local neural network \cite{nonlocal} which is based on I3D \cite{i3d} for video modelling and leverages the spatial-temporal nonlocal relationships therein. However, 3D CNN is computational costly and training 3D CNN models inflated from deeper network suffers from performance drop due to batch size reduction.

\begin{figure*}[!ht]
\begin{center}
\includegraphics[width=0.9\textwidth]{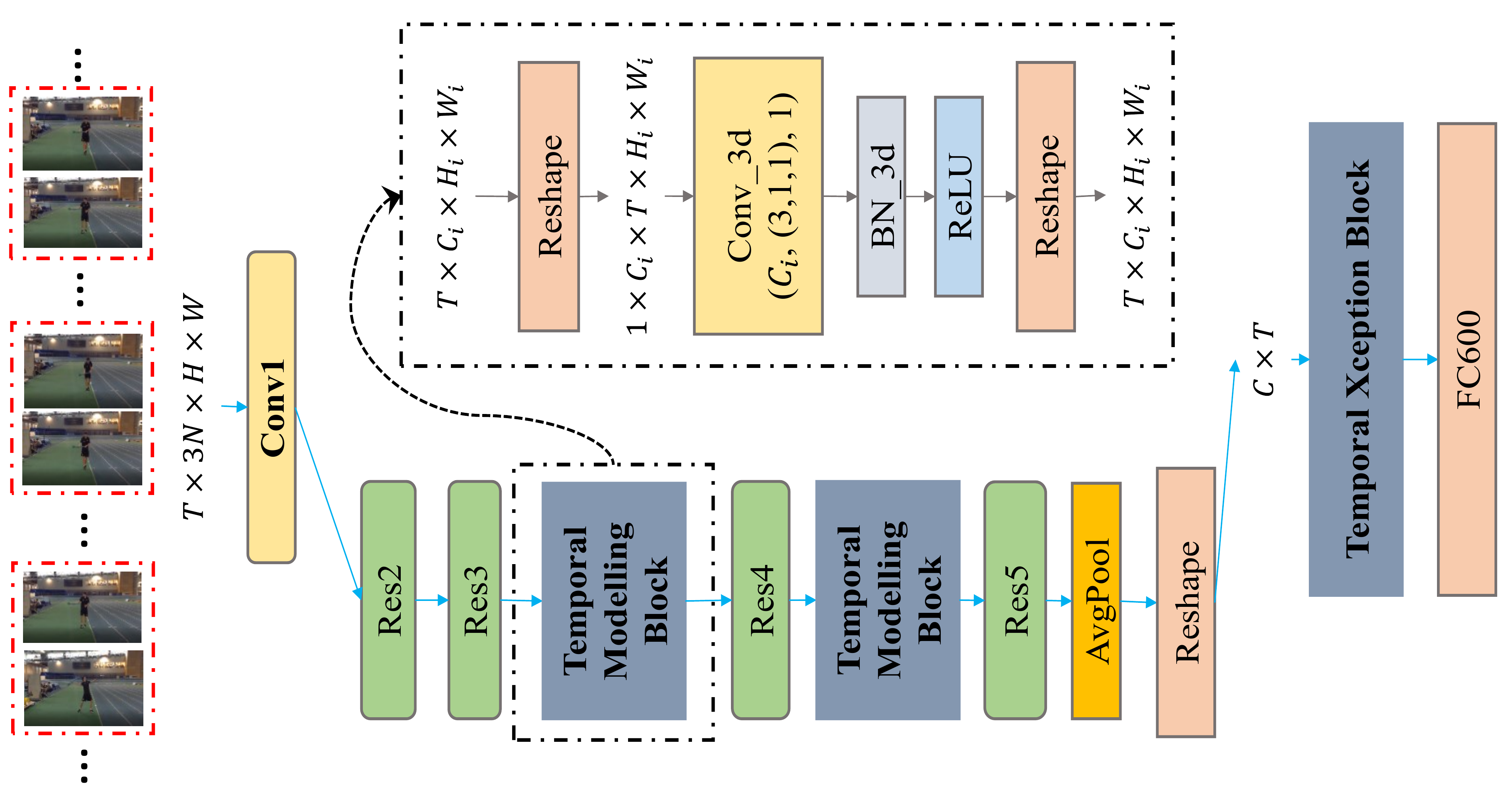}
\end{center}
   \caption{Illustration of constructing StNet based on ResNet \cite{resnet} backbone. The input to StNet is a $T\times 3N \times H \times W$ tensor. Local spatial-temporal patterns are modelled via 2D Convolution. 3D convolutions are inserted right after the Res3 and Res4 blocks for long term temporal dynamics modelling. The setting of 3D convolution (\# Output Channel, (temporal kernel size, height kernel size, width kernel size), \# groups) is ($C_i$, (3,1,1), 1). }
\label{fig:stnet}
\end{figure*}

In this challenge, we propose a novel framework called Spatial-temporal Network (StNet) to jointly model spatial-temporal correlations for video understanding. StNet first models local spatial-temporal correlation by applying 2D convolution over a $3N$-channel \emph{super image} which is formed by sampling $N$ successive RGB frames from a video and concatenating them in the the channel dimension. As for long range temporal dynamics, StNet treats 2D feature maps of uniformly sampled $T$ \emph{super images} as 3D feature maps whose temporal dimension is $T$ and relies on 3D convolution with temporal kernel size of 3 and spatial kernel size of 1 to capture long range temporal dependency. With our proposed StNet, both local spatial-temporal relationship and long range temporal dynamics can be modelled in an end-to-end fashion. In addition, large number of convolution kernel parameters is avoided because we can model local spatial-temporal with 2D convolution and spatial kernel size of 3D convolution in StNet is set to 1.

Video source contains such multi-modal information as appearance information in the RGB frames, motion information among successive video frames and acoustic information in its audio signal. Existing works have proved that fusing multi-modal information is helpful \cite{tsn, attentioncluster, kinetics400_win}. In this challenge, we also utilize multiple modalities to boost the recognition performance. We improve our formerly proposed temporal Xception network \cite{kinetics400_win} and enable
it to integrate both early-fusion and later-fusion features of multi-modal information. This model is referred to as improved temporal Xception network (iTXN) in the following .

\section{Spatial-Temporal Modelling}
\begin{figure*}[!t]
\begin{center}
\includegraphics[width=0.95\textwidth]{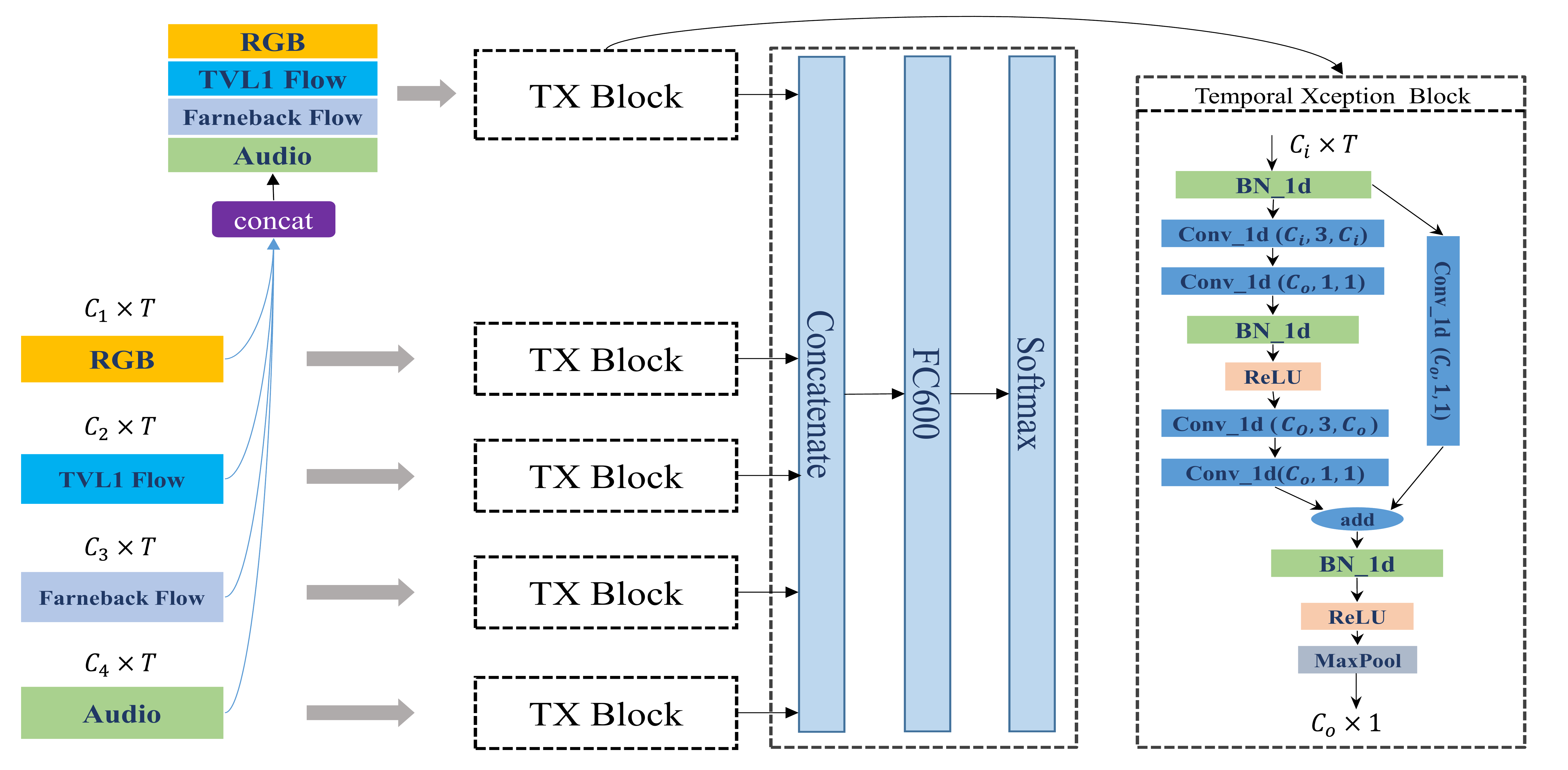}
\end{center}
   \caption{Block diagram of our proposed improved temporal Xception network (iTXN) framework for multi-modality integration. It is built based upon the temporal Xception network \cite{kinetics400_win}. RGB, TV\_L1 flow, Farneback flow and audio feature sequences are encoded individually for later-fusion and encoded jointly for early fusion with temporal Xception block, respectively. Numbers in bracket of temporal Xception block denote (\# Output Channel, kernel size, \# group) of Conv\_1d layer.}
\label{fig:iTXN}
\end{figure*}
The proposed StNet can be constructed from existing state-of-the-art 2D CNN frameworks, such as ResNet \cite{resnet}, Inception-Resnet \cite{inceptionresnet} and so on. Taking ResNet as example, Fig.\ref{fig:stnet} illustrates how we can build StNet. Similar to TSN \cite{tsn}, we choose to model long range temporal dynamics by temporal snippets sampling rather than inputing the whole video sequence. One of the differences from TSN is that we sample $T$ temporal segments which consists of $N$ contiguous RGB frames rather than one single frame. These $N$ frames are stacked to form a \emph{super image} whose channel size is $3N$, so the input to the network is a tensor of size $T\times 3N \times H \times W$. We choose to insert two temporal modelling blocks right after the Res3 and Res4 block. The temporal modelling blocks are designed to capture the long-range temporal dynamics inside a video sequence and they can be implemented easily by leveraging Conv3d-BN3d-ReLU. Note that existing 2D CNN framework is powerful enough for spatial modelling, so we set both height kernel size and width kernel size of 3D convolution as 1 to save model parameters while the temporal kernel size is empirically set to be 3. As an augmentation, we append a temporal Xception block \cite{kinetics400_win} to the global average pooling layer for further temporal modelling. Details about temporal Xception block can be found in the most right block of Fig.\ref{fig:iTXN}.

To build StNet from other 2D CNN frameworks such as InceptionResnet V2 \cite{inceptionresnet}, ResNeXt \cite{resnext} and SENet \cite{senet} is quite similar to what we have done with ResNet, therefore, we do not elaborate all such details here. In our current setting, $N$ is set to 5, $T$ is 7 in the training phase and $25$ in the testing phase. As can be seen, StNet is an end-to-end framework for joint spatial-temporal modelling. A large majority of its parameters can be initialized from its 2D CNN counterpart.
The initialization of the rest parameters following the below rules: 1) weights of Conv1 can be initialized following what the authors have done in I3D \cite{i3d}; 2) parameters of 1D or 3D BatchNorm layers are initialized to be identity mapping; 3) biases of 1D or 3D Conv are initially set to be zeros and weights are all set to $1/(3\times C_i)$, where $C_i$ is input channel size.

\section{Multi-Modal Fusion}
Videos consist of multiple modalities. For instances, appearance information is contained in RGB frames, motion information is implicitly shown by the gradually change of frames along time and audio can provide acoustic information. For a video recognition system, utilizing such multi-modal information effectively is beneficial for performance improvement. Existing works \cite{attentioncluster, kinetics400_win} have evidenced this point.

In this piece of work, we also follow the common practice to boost our recognition performance by integrating multi-modal information, i.e., appearance, motion and audio. Appearance can be explored from RGB frames with existing 2D/3D solution as well as our proposed StNet. To better utilize motion information, we extract optical flows from video sequences not only with the TV\_L1 algorithm \cite{tvl1} but also with the Farneback algorithm \cite{farneback}. As for audio information, we simply follow what have been done in \cite{audiovgg, kinetics400_win}.

Fusing multi-modal information have been extensively researched in the literature. Early-fusion and later-fusion are the most common methods. In this paper, we propose to combine early-fusion and later-fusion in one single framework. As is shown in Fig.\ref{fig:iTXN}, pre-extracted features of RGB, TV\_L1 flow, Farneback flow and audio are concatenated along with the channel dimension and its output is fed into a temporal Xception block for early fusion. These four feature modalities are also encoded with temporal Xception block individually. Afterwards, the early-fusion feature vector are concatenated with the individually encoded features of the four modality for classification.

\section{Experiments}
In this section, we report some experimental results to verify the effectiveness of our proposed frameworks. All the base RGB, flow and audio models evaluated in the following subsection are pre-trained on the Kinetics-400 training set and finetuned on the Kinetics-600 training set. All the results reported below are evaluated on the Kinetics-600 validation set.
\subsection{Spatial-Temporal Modelling}
To show the effectiveness of the proposed StNet, we have trained StNet with InceptionResnet V2 \cite{inceptionresnet} and SE-ResNeXt 101 \cite{senet, resnext} and a series of baseline RGB models, denoted as StNet-IRv2 and StNet-se101 respectively. As we know, the state-of-the-art 2D CNN models for action recognition is TSN \cite{tsn}, and we implemented TSN with InceptionResnet V2 and SE-ResNeXt 152 backbone networks. In the following context, we denote these two models as TSN-IRv2 and TSN-se152 respectively. We also introduced VLAD encoding + SVM on the TSN-IRv2 Conv2d\_7b feature. Nonlocal neural network is state-of-the-art 3D CNN model for video classification, so we also finetuned nonlocal-Res50-I3D net as a baseline model with the codes released by the authors. Due to the time limitation, we cannot afford training such big model as nonlocal-Res101-I3D.

\begin{table}[!h]
\centering
\caption{Performance comparison among StNet and baseline RGB models. }
\label{t:snippets}
\begin{tabular}{ c | c  }
  \hline\hline		
  Model & Prec@1 \\
  \hline
   TSN-IRv2 (T=50, cropsize=331)& 76.16\%\\
   TSN-se152 (T=50, cropsize=256) & 76.22\%\\
   TSN-IRv2 + VLAD + SVM & 75.6\%\\
   \hline
  Nonlocal Res50-I3D (1crop of 32 frames) & 71.1\% \\
  Nonlocal Res50-I3D (30crops) & 78.6\%\\
  \hline
  StNet-se101 (T=25, cropsize=256) & 76.08\% \\
  StNet-IRv2 (T=25, cropsize=331)& \textbf{78.99\%}\\
  \hline
  \hline
\end{tabular}
\end{table}

Evaluation results are presented in Tabel.\ref{t:snippets}. We can see from this table that StNet-IRv2 outperforms TSN-IRv2 by up to 2.83\% in top-1 precision and it also achieves better performance than nonlocal-Res50-I3D net. Please note that our StNet-se101 performs comparable with TSN-se152, which also evidences the superiority of the StNet framework.

\subsection{Multi-Modal Fusion}
In this work, we exploit not only RGB information, but also TV\_L1 flow \cite{tvl1}, Farneback flow \cite{farneback} and audio information \cite{audiovgg} extracted from video sources. The recognition performances with each individual modality are listed in Table.\ref{t:m}. For multi-modality fusion, StNet-IRv2 RGB feature, TSN-IRv2 TV\_L1 flow feature, TSN-se152 Farneback flow feature and TSN-VGG audio feature are used for better complementarity.

\begin{table}[!h]
\centering
\caption{Recognition performance of each individual modality. }
\label{t:m}
\begin{tabular}{ c | c  }
  \hline\hline		
  Modality & Prec@1 \\
  \hline
   TSN-IRv2 TV\_L1 & 65.1\%\\
  \hline
   TSN-IRv2 Farneback flow & 69.3\%\\
   TSN-se152 Farneback flow & 71.3\%\\
  \hline
  StNet-IRv2 RGB & 78.99\\
  \hline
  TSN-VGG audio & 23\% \\
  \hline
  \hline
\end{tabular}
\end{table}

To evaluate iTXN which is designed for multi-modal fusion, we compared it with several baselines: AttentionClusters \cite{attentioncluster}, Fast-Forward LSTM and temporal Xception network which are proposed in \cite{kinetics400_win}. The results are shown in the Table.\ref{t:fusion}. From this table, we can see that  iTXN is a good framework for integrating multiple modalities.

Our final results are obtained by ensembling multiple single modality models and several multi-modal models by gradient boosting decision tree (GBDT) \cite{gbdt}. After model ensemble, we finally achieve top-1 and top-5 precision of 85.0\% and 96.9\% on the validation set.
\begin{table}[!h]
\centering
\caption{Recognition performance of multi-modal fusion and model ensemble. }
\label{t:fusion}
\begin{tabular}{ c | c | c }
  \hline\hline		
  Model & Prec@1 & Prec@5\\
  \hline
   temporal Xception network &81.8\% &95.6\% \\
   Fast-Forward LSTM &81.6\% &95.1\% \\
   AttentionClusters &82.3\% &\textbf{96.0\%} \\
  \hline
  iTXN &\textbf{82.4\%} &95.8\% \\
  \hline
  Model Ensemble &\textbf{85.0\%} & \textbf{96.9\%} \\
  \hline
  \hline
\end{tabular}
\end{table}

\section{Conclusion}
In this challenge, we proposed a novel StNet end-to-end framework to jointly model spatial-temporal patterns in videos for human action recognition. In order to better integrate multi-modal information which is naturally contained in video sources, we improved temporal Xception network to combines both early-fusion and later-fusion of multiple modalities. Experiment results have evidenced the effectiveness of the proposed StNet and iTXN.

%\section{Acknowledgement}
%We thank our colleague Xin Li for giving us hands on data preprocessing and Bo Cheng for his study on VLAD baseline. We also appreciate Xiao Liu for valuable discussions with him.

\bibliographystyle{splncs}
%\bibliography{egbib}

\end{document}